\pdfoutput=1
\documentclass[11pt]{article}

\usepackage[final]{acl}

\usepackage{times}
\usepackage{latexsym}

\usepackage[T1]{fontenc}
\usepackage[utf8]{inputenc}

\usepackage{microtype}

\usepackage{inconsolata}

\usepackage{bm}
\usepackage{tabularx}
\usepackage{graphicx}
\usepackage{color}
\usepackage{amsmath}
\usepackage{colortbl}

\title{Automated Essay Scoring Using Grammatical Variety and Errors \\ with Multi-Task Learning and Item Response Theory}

\author{Kosuke Doi \quad Katsuhito Sudoh \quad Satoshi Nakamura \\
  Nara Institute of Science and Technology \\
  \texttt{\{doi.kosuke.de8, sudoh, s-nakamura\}@is.naist.jp} \\}
\begin{document}
\maketitle
\begin{abstract}
This study examines the effect of grammatical features in automatic essay scoring (AES).
We use two kinds of grammatical features as input to an AES model: (1)~grammatical items that writers used correctly in essays, and (2)~the number of grammatical errors. 
Experimental results show that grammatical features improve the performance of AES models that predict the holistic scores of essays.
Multi-task learning with the holistic and grammar scores, alongside using grammatical features, resulted in a larger improvement in model performance.
We also show that a model using grammar abilities estimated using Item Response Theory (IRT) as the labels for the auxiliary task achieved comparable performance to when we used grammar scores assigned by human raters.
In addition, we weight the grammatical features using IRT to consider the difficulty of grammatical items and writers' grammar abilities.
We found that weighting grammatical features with the difficulty led to further improvement in performance.\footnote{The code is publicly available at \url{https://github.com/ahclab/aes-grammar-mtl-irt}.}
\end{abstract}

%%%%%%

\section{Introduction}

Automated Essay Scoring (AES) is a task that automatically grades essays.
Essay assignments are widely used in language tests and classrooms to assess learners' writing abilities, while grading them takes time and effort for human raters.
Maintaining inter- and intra-rater reliability is another issue associated with human scoring.
AES can help alleviate these problems and has been attracting more attention in recent years.

The grading methods for essays can be roughly categorized into two types: holistic scoring and analytic scoring.
The former assigns a single score to an essay based on its overall performance, while the latter assigns different scores to various aspects of the essay, such as grammar, vocabulary, content, or organization \citep{Weigle_2002}.
However, rubrics for holistic scoring typically contain descriptions of several aspects of writing used in analytic scoring (\emph{e.g.}, TOEFL iBT Independent Writing Rubric).

Among those aspects, we focus on grammatical features, inspired by the research on criterial features for the levels of the Common European Frameworks of References (CEFR) \citep{Council_of_Europe2001_cefr} in L2 English \citep{Hawkins_2012}.
The CEFR, one of the influential frameworks in language teaching, describes language abilities in functional terms (\emph{i.e.}, can-do statements, such as ``Can write short, simple essays on topics of interest'').
However, it is grammatical items and lexis that realize the functions written in can-do statements.
To fully develop and elaborate their ideas in essays, they need to use a wide range of grammatical items.
In fact, grammar plays an important role in essay scoring.
Researches on writing in the second language acquisition field have been focusing on syntactic complexity\footnote{Syntactic complexity refers to the extent to which a learner can use a wide variety of both basic and sophisticated structures \citep{Wolfe-Quintero_1998}.} and accuracy (see \citealp{Kuiken_2023, housen-et-al-2012}).

\citet{Hawkins_2012} identified grammatical items that learners at a certain level and higher can use correctly and items that learners at a certain level are prone to making mistakes in.
It is known that human raters look for those features consciously or unconsciously when they evaluate learners' performance, and explicit use of grammatical features in AES will improve model performance.

Grammatical features have been used in many feature-engineering AES models (see  \citealp{Ke2019_survey}) as well as in hybrid models, which incorporate handcrafted features into deep neural network AES models \citep{dasgupta-etal-2018-augmenting,uto-etal-2020-neural,banno-matassoni-2022-cross}.
In \citet{yannakoudakis-etal-2011-new}, features representing grammatical structures were used together with other linguistic features.
However, in many previous studies (\emph{e.g.}, \citealp{Vajjala2018_automated,uto-etal-2020-neural}), grammatical items used correctly were aggregated into measures of grammatical complexity (\emph{e.g.},~ratio of dependent clauses per clauses; see \citealp{Wolfe-Quintero_1998}) rather than individual grammatical items (\emph{e.g.},~adverbial clause \emph{if}, adverbial clause \emph{so that}) even though the difficulties of individual grammatical items are different.

In this paper, we propose to use individual grammatical items as inputs to hybrid AES models that predict holistic scores, and leverage the models to incorporate the variety of grammatical items in grading essays.
We also use frequencies of grammatical errors corrected by a modern grammatical error correction model (GECToR-large; \citealp{tarnavskyi-etal-2022-ensembling}) as model inputs.
The grammatical features are combined with an essay representation and passed into a fully connected feed-forward neural network to predict the score of an input essay.
Our models used BERT \citep{devlin-etal-2019-bert} to learn essay representations following the current state-of-the-art AES models \citep{yang-etal-2020-enhancing,cao-2020-domain,wang-etal-2022-use}.

To utilize grammatical features more effectively, we develop a multi-task learning framework that jointly learns to predict holistic scores and grammar scores of essays.
We use two types of grammar scores: (1)~scores assigned to essays by human raters and (2)~writers' latent abilities estimated based on patterns of grammar usage using Item Response Theory (IRT) \citep{Lord-1980}.
Note that teacher labels are not necessary for estimating the latent abilities using IRT.

IRT estimates not only each writer's ability but also the characteristics of each item (\emph{i.e.}, individual grammatical item), such as discrimination and difficulty parameters.
Therefore, we use these IRT parameters to weight grammatical items (\emph{e.g.}, award writers who use a difficult grammatical item; see Section~\ref{sssec:weighted-pf}).

In summary, the contributions of this paper are as follows:
\begin{itemize}
    \item We propose to use individual grammatical items and grammatical errors as inputs to AES models, and leverage the models to consider grammar use in predicting holistic scores of essays.
    \item We develop a multi-task learning framework that jointly learns to predict holistic scores and grammar scores of essays.
    \item We apply IRT to writers' grammar usage patterns and (1)~use estimated latent abilities for multi-task learning, and (2)~use IRT parameters to weight grammatical items when we feed them to AES models.
    \item We show the effectiveness of incorporating grammatical features into BERT-hybrid AES models. Our method shows a significant advantage on some essay assignments in the Automated Student Assessment Prize (ASAP) dataset\footnote{https://www.kaggle.com/c/asap-aes}.
\end{itemize}

%%%%%%

\section{Related Work}

\subsection{Automated Essay Scoring}

Early AES models predict essay scores using handcrafted features (see \citealp{Ke2019_survey}).
For example, e-rater \cite{Burstein2004_e-rater} uses 12 features, including grammatical errors and lexical complexity measures.
\citet{yannakoudakis-etal-2011-new} automatically extracted various linguistic features, including grammatical structures, using a parser.
These features were weighted and used to train SVM ranking models.
\citet{Vajjala2018_automated} reported that measures of grammatical complexity and errors were assigned large weights among various linguistic features.

Recently, a deep neural network-based approach has become popular.
AES models based on RNN \citep{taghipour-ng-2016-neural}, Bi-LSTM \cite{alikaniotis-etal-2016-automatic}, and pre-trained language models \citep{nadeem-etal-2019-automated,yang-etal-2020-enhancing,cao-2020-domain,wang-etal-2022-use} have been proposed.
In addition, a hybrid model, which incorporates handcrafted features into a deep neural network-based model, has been proposed \citep{dasgupta-etal-2018-augmenting,uto-etal-2020-neural,banno-matassoni-2022-cross}.

AES using a large language model has also been explored.
\citet{mizumoto-2023} demonstrated that using linguistic features in GPT-3 improved AES performance.
\citet{yancey-etal-2023-rating} reported that providing a small number of scoring examples to GPT-4 led to comparable performance to models trained on hundreds of thousands of data based on 85 language features.

This study examines the effect of explicitly considering grammatical features in a hybrid AES model by incorporating individual grammatical items as model inputs and weighting them using IRT parameters.

\subsection{Multi-Task Learning}

Multi-task learning (MTL) \citep{Caruana-1997} is a method that improves the generalization performance of the main task by training a single model to perform multiple tasks simultaneously.
MTL has been used in previous studies in AES, and shown to be effective.
\citet{cummins-etal-2016-constrained} used MTL to overcome the lack of task-specific data in the ASAP dataset by treating each essay prompt as a different task.
\citet{Xue-2021-hierarchical} also trained a model jointly on eight different prompts in the ASAP dataset using BERT.

There are also studies that have performed MTL with other NLP tasks.
\citet{Cummins2018_multi-task} trained an LSTM jointly on grammatical error detection and AES.
While the error detection task in \citet{Cummins2018_multi-task} required the model to predict whether a particular token was errorful, ones in \citet{elks-2021-using-transfer} require to (1)~predict a sentence contains errors or (2)~classify tokens by a type of error (\emph{e.g.}, correct, lexical, form).
Other auxiliary tasks used in previous studies include morpho-syntactic labeling, language modeling, and native language identification \citep{craighead-etal-2020-investigating}, sentiment analysis \citep{muangkammuen-fukumoto-2020-multi}, predicting the level of each token \citep{elks-2021-using-transfer}, and predicting span, type, and quality of argumentative elements \citep{ding-etal-2023-score}. 

In this paper, we train models jointly on holistic scores and grammar scores.
This is similar to AES models that predict multiple essay traits simultaneously \citep{mathias-bhattacharyya-2020-neural,Hussein-2020,mim-etal-2019-unsupervised,Ridley_He_Dai_Huang_Chen_2021}, but the difference between them and ours is that we explicitly incorporate grammatical features to a model, which are related to the score to be predicted.

\subsection{Item Response Theory}

IRT is a probabilistic model that has been widely used in psychological and educational measurement \citep{Hambleton-1991-irt}.
An IRT model expresses the probability of a correct response to a test item as a function of the item parameters, which represent the characteristics of the item, and the ability parameter, which represents the ability of the examinee.

Previous studies in AES used IRT to mitigate raters' bias \citep{Uto-2021-learning}, integrate prediction scores from various AES models \citep{Aomi-2021,Uto-2023-integration}, and predict multiple essay traits \citep{Uto-2021-multidimensional,shibata-uto-2022-analytic}.
These studies employed a multidimensional IRT model since unidimensionality cannot be assumed for the subject to which IRT is applied.

In contrast, we regard individual grammatical items as test items, assuming that whether grammar items are used correctly constitutes grammar ability (\emph{i.e.}, satisfy the assumption of unidimensionality). 
We model writers' grammar ability using two-parameter logistic model \citep{Lord-1952}, formulated by the following equation:
\begin{equation}
  P_{ij} (\theta_i) = \frac{1}{1 + \exp(-Da_j(\theta_i - b_j))}
\end{equation}
where $P_{ij} (\theta_i)$ is the probability that the writer~$i$ with ability~$\theta_i$ uses the grammatical item~$j$ correctly, $a_j$ is the discrimination parameter for item~$j$, and $b_j$ is the difficulty parameter for item~$j$.
$D$ is a scaling factor and set to $1.0$ in this paper.

%%%%%%

\section{Proposed Method}

\subsection{Grammatical Features}
\label{ssec:grammatical-features}

The Common European Frameworks of References (CEFR) \citep{Council_of_Europe2001_cefr} is an internationally recognized framework for language proficiency.
It divides proficiency into six levels ranging from A1 (beginner) to C2 (advanced).
Due to the language-neutral nature of the CEFR, what grammatical and lexical properties learners develop across the CEFR levels has been studied language by language.

Such properties (criterial features) in English have been identified by English Profile Programme \citep{Hawkins_2012}.
Criterial features refer to linguistic properties that are characteristic and indicative of L2 proficiency levels and that distinguish higher levels from lower (\emph{ibid}).
They identified positive linguistic features (PFs; grammatical items that learners can use correctly at a certain level and higher) and negative linguistic features (NFs; grammatical items that learners at a certain level are prone to making mistakes in) in relation to the CEFR levels.

Based on the analyses of human raters' grading performance in actual exams, \citet{hawkins-buttery-2009} have argued that they develop clear intuitions about these properties.
We expect that allowing a model to learn grammar representations using grammatical features would improve the AES performance.
Table \ref{tab:g_features} shows PFs and NFs used in our experiments.
The following sections describe them in detail.

\begin{table}[t]
\centering
\small
\tabcolsep 3pt
%\begin{tabular}{ll}
\begin{tabularx}{\linewidth}{lX}
\hline
Features & Descriptions \\
\hline
type256 & 256 grammatical items, whether a writer use the items \\
\hline
err54 & 54 types of errors, \# of errors \\
\hline
multiply\_b & Modify \texttt{type256} with item difficulty \\
prob & Replace elements in \texttt{type256} with the probabilities of using the items correctly\\
multiply\_prob & Weight \texttt{type256} with the probabilities \\
add\_prob & Consider both the actual use (\texttt{type256}) and the probabilities \\
\hline
%\end{tabular}
\end{tabularx}
\caption{Grammatical features used in our experiments. The number of errors is relative freq. per 100 words.}
\label{tab:g_features}
\end{table}

\subsubsection{Positive Linguistic Features}

PFs were extracted using a toolkit for frequency analysis of grammatical items, which is provided by the CEFR-J Grammar Profile \citep{Ishii-2018}.
It extracts 501 grammatical items in text based on regular expressions and calculates the frequencies of them.
We converted the frequencies into the 256-dimensional vector (\texttt{type256}) based on CEFR-J Grammar Profile for Teachers\footnote{https://www.cefr-j.org/download.html\#cefrj\_grammar \\ The toolkit distinguishes the same items in different sentence types such as the affirmative or negative, while CEFR-J Grammar Profile for Teachers does not.} as \mbox{$\bm{g_i} =\{g_{i1}, g_{i2}, ..., g_{i256}\}$}.
Each dimension corresponds to a grammatical item, and $g_{ij} = 1$ if the writer~$i$ used the item~$j$ in the essay, and 0 if not.

\subsubsection{PFs Weighted using IRT Parameters}
\label{sssec:weighted-pf}

Researches on criterial features have shown that learners master more and more grammatical items across the CEFR levels, but \texttt{type256} does not consider the difficulties of the items.
Therefore, we weight them using the IRT parameters.

%Let $g_{nj}$ represent whether the writer $n$ uses the grammatical item $j$ in the essay.
We transform $g_{ij}$ in the following four ways:
\begin{description}
    \setlength{\itemsep}{-1pt}
    \setlength{\leftskip}{0.3cm}
    \item[multiply\_b:] $g'_{ij} = g_{ij} \times b_j$
    \item[prob:] $g'_{ij} = P_{ij}(\hat{\theta}_i)$
    \item[multiply\_prob:] $g'_{ij} = g_{ij} \times P_{ij}(\hat{\theta}_i)$
    \item[add\_prob:] $g'_{ij} = \alpha g_{ij} + (1-\alpha) P_{ij}(\hat{\theta}_i)$
\end{description}
where $\hat{\theta}_i$ is the grammatical ability of the writer~$i$ estimated based on the patterns of grammar usage using IRT, and $\alpha$ is a weighting parameter.
$\alpha$ was set to $0.5$ in this paper.

\texttt{multply\_b} aims to consider the difficulty of items by multiplying the difficulty parameter for the item.
However, writers might not have used some grammatical items because of the essay topic although they had enough abilities to do.
Therefore, we use $P_{ij}(\hat{\theta}_i)$, which shows the probability that the writer~$i$ with ability $\hat{\theta}_i$ can use the item~$j$ correctly.
In \texttt{prob} $g_{ij}$ is replaced with $P_{ij}(\hat{\theta}_i)$, while in \texttt{multiply\_prob} and \texttt{add\_prob} the two values are combined to consider both the ability of writers and the actual use in essays.

IRT parameters were estimated independently from the prediction of essay scores.
The IRT parameters were frozen during the training of scoring models.

\subsubsection{Negative Linguistic Features}

We calculated the number of grammatical errors per 100 words as NFs.
Specifically, we created the 54-dimensional vector (\texttt{err54}) based on error tags assigned by ERRANT \citep{bryant-etal-2017-automatic}\footnote{Based on all possible combinations of the error types and categories. We tried the 24-dimensional vector, which was based on the error types, but the 54-dimensional vector improved the model performance more.}.
We used GECToR-large \citep{tarnavskyi-etal-2022-ensembling} to correct grammatical errors in essays.

\subsection{Model Architecture}
\label{ssec:model-architecture}

\begin{figure}[t]
\centering
\includegraphics[width=0.99\linewidth]{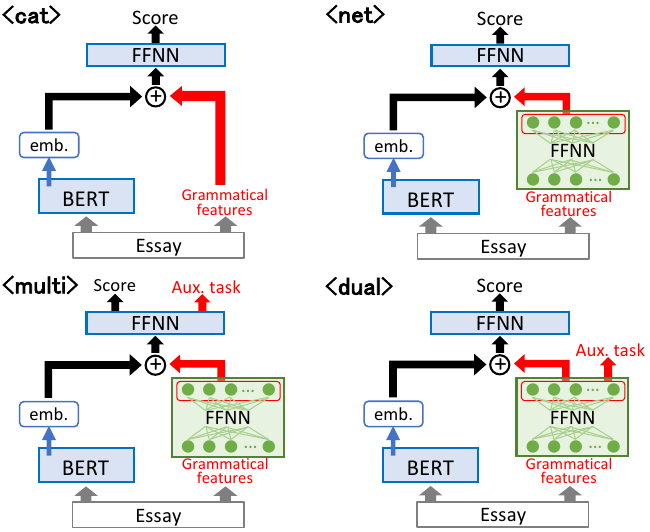}
\caption{The architectures of proposed models}
\label{fig:model}
\end{figure}

Our model takes a batch of essays and grammatical features as input and predicts the holistic scores of the essays.
We prepare a model that takes only a batch of essays as input for a baseline.
Essay representations are obtained from the \texttt{[CLS]} token of the BERT model.

Grammatical features are used in the four settings shown in Figure~\ref{fig:model}.
In \texttt{cat}, we concatenate the essay representation and the vector of grammatical features, and feed it to a fully connected feed-forward neural network (FFNN).
In \texttt{net}, we first feed the vector of grammatical features to an FFNN and concatenate the representation from the final layer with the essay representation.
In \texttt{multi}, we perform multi-task learning with the model architecture of \texttt{net}.
The FFNN in \texttt{multi} consists of shared layers only, and does not have task-specific layers\footnote{We tried models with task-specific layers, but the performance was worse than ones without them.}.
In \texttt{dual}, the predicted values for the auxiliary task are output from the FFNN for grammatical features.

As the labels for the auxiliary task in \texttt{multi} and \texttt{dual},  we used grammar scores assigned to essays by human raters, which is available in ASAP and ASAP++ dataset \citep{mathias-bhattacharyya-2018-asap}, and grammar abilities estimated using IRT.
Grammar abilities can be estimated from writers' grammar usage patterns without any teacher labels.

\section{Experiments}

\subsection{Data and Evaluation}

We used the ASAP and the ASAP++ dataset in our experiments.
The ASAP consists of essays for eight different prompts, with holistic scores for Prompts 1-6 and analytic scores for Prompts 7-8.
In Prompt 7 and 8, the weighted sum of the analytic scores constitutes the total score, which is the target of prediction by our models.
ASAP++ includes analytic scores of essays for Prompt 1-6.
We developed AES models that predict the holistic score for each essay prompt.
From analytic scores, we only used ones related to grammar\footnote{Conventions for Prompt 1, 2, 7, and 8. Language for Prompt 3-6.}.

We evaluated the scoring performance of our models using the Quadratic Weighted Kappa (QWK) on the ASAP dataset.
Following the previous studies, we adopted 5-fold cross validation with 60/20/20 split for train, development, and test sets, which was provided by \citet{taghipour-ng-2016-neural}.

\subsection{Settings}

As explained in Section~\ref{ssec:model-architecture}, we developed our AES models based on BERT.
We used \texttt{bert-base-uncased} provided by Hugging Face\footnote{https://github.com/huggingface/transformers}.
The maximum input length was set to 512.

We normalized essay scores in the range of \mbox{$[-1, 1]$}.
The mean squared error (MSE) loss was employed for both the main and auxiliary tasks.
We updated the parameters for the FFNN and the BERT layers.
The number of hidden layers in the FFNN for grammatical features was set to 3, and the number of the nodes in the hidden layer to one-half the dimension of the grammatical features.
The number of hidden layers in the FFNN on the top was set to $\{1, 2, 3, 4, 5, 7, 10\}$ for \texttt{cat}, $\{1, 2, 3\}$ for \texttt{net}, and $\{2, 3\}$ for \texttt{multi} and \texttt{dual} and we chose the value that achieved the best QWK score on the development set for Prompt~1.
The number of the nodes was set to 512.
For both FFNNs, we adopted relu as the activation function and set the dropout ratio to 0.2.
In \texttt{multi} and \texttt{dual}, we tried $\{0.8, 0.6\}$ for the weights of the loss function for the main task.
We used Adam optimizer \citep{kingma-2015} with a learning rate of 1e-5.
We trained models with the batch size $\{4, 8, 16, 32\}$ for 10 epochs.
In the following sections, we report the scores on test sets for the batch size with the highest QWK on the development set for each essay prompt.
The scores are the average of three experiments with different seed values.

\section{Results}

\subsection{Hyperparameters for Each Model Architecture}
\label{ssec:hyperparameters-for-models}

\begin{table}[t]
\centering
\small
\tabcolsep 3pt
\begin{tabular}{lccccccc}
\hline
 & \multicolumn{7}{l}{\# of hidden layers} \\
\cline{2-8}
Model & 1 & 2 & 3 & 4 & 5 & 7 & 10 \\
\hline
baseline & .813 & -- & -- & -- & -- & -- & -- \\
\hline
cat & .792 & \textbf{.825} & .814 & .813 & .801 & .766 & .722 \\
\hline
net & .812 & \textbf{.824} & .817 & -- & -- & -- & -- \\
\hline
multi--hum (0.8) & -- & .819 & \textbf{.827} & -- & -- & -- & -- \\
multi--hum (0.6) & -- & .804 & .812 & -- & -- & -- & -- \\
\hline
dual--hum (0.8) & -- & .816 & \textbf{.824} & -- & -- & -- & -- \\
dual--hum (0.6) & -- & .820 & .819 & -- & -- & -- & -- \\
\hline
\end{tabular}
\caption{Comparison of the number of hidden layers in FFNN on the top (type256, Prompt~1, QWK dev)}
\label{tab:compare-num-hidden}
\end{table}

Using \texttt{type256} for the grammar features, we searched for the optimal hyperparameters for each model architecture.
Table~\ref{tab:compare-num-hidden} shows the QWK results on the development set of Prompt~1 when we changed the number of hidden layers in the FFNN on the top.
When the number of hidden layers was set to 1 in \texttt{cat}, QWK was lower than the baseline (.792 vs. .813).
QWK became the highest when the number of hidden layers was set to 2, while it got lower as the number of hidden layers increased.
In \texttt{net}, the architecture with 2 hidden layers achieved the highest QWK.
In both \texttt{multi--hum} and \texttt{dual--hum}\footnote{``--hum'' represents that grammar scores assigned by human raters were used. ``--irt'' is added when grammar abilities estimated using IRT are used.}, QWK became the highest when the number of hidden layers was set to 3 and the weight of the loss for the main task to 0.8.
In the subsequent experiments, we trained models using these hyperparameters.

\subsection{Comparison among Model Architectures}
\label{ssec:compare-model}

\begin{table}[t]
\centering
\small
\tabcolsep 2pt
\begin{tabular}{lccccccccc}
\hline
 & \multicolumn{8}{l}{Prompt} & \\
\cline{2-10}
Model & 1 & 2 & 3 & 4 & 5 & 6 & 7 & 8 & avg. \\
\hline
baseline & .799 & .662 & .662 & .804 & .801 & .809 & .821 & .726 & .760 \\
\hline
+ type256 & & & & & & & & & \\
%cat   & .818 & .675 & .673 & .819 & .806 & .808 & .832 & .734 & .771 \\
cat & .819 & .674 & .675 & .801 & .809 & .809 & .830 & .721 & .767 \\
%net   & \textbf{.822} & \textbf{.684} & .685 & .811 & .804 & .813 & .834 & .746 & .775 \\
net & .814 & .679 & .678 & .810 & .806 & .806 & .831 & .737 & .770 \\
%multi & .812 & .682 & .694 & .817 & .808 & \textbf{.814} & \textbf{.837} & .749 & .777 \\
multi--hum & .816 & .678 & .683 & .812 & .810 & .811 & .830 & .746 & .773 \\
%dual  & .820 & .675 & \textbf{.700} & \textbf{.820} & \textbf{.809} & .806 & .831 & \textbf{.763} & \textbf{.778} \\
dual--hum & .818 & .673 & .687 & .819 & .807 & .813 & .833 & .750 & .\textbf{775} \\
\hline
\end{tabular}
\caption{Comparison among model architectures (type256, QWK test)}
\label{tab:compare-models}
\end{table}

Using \texttt{type256} for the grammar features, we compared the model performance among the four model architectures.
Table~\ref{tab:compare-models} shows the QWK results on the test set of all prompts.
By using \texttt{type256}, the average QWK score for all essays improved in all proposed models, compared to the baseline (See avg. in Table~\ref{tab:compare-models}).

In \texttt{cat}, however, the QWK scores did not improve in three prompts (Prompt~4, 6, and 8), which suggests that simple concatenation of essay representations and grammatical features was not sufficient enough to take advantage of the information that the grammatical features have.
In \texttt{net}, only Prompt~6 did not improve from the baseline, and it seems effective to feed the grammatical features to an FFNN before concatenating with essay representations.

The QWK scores for the models with the auxiliary task (\texttt{multi--hum} and \texttt{dual--hum}) were higher than the others.
Even when looking at the QWK scores for each essay prompt individually, the scores improved for all prompts.
These results suggest that multi-task learning with grammar scores is effective to take advantage of grammatical features.

\texttt{Dual--hum} achieved the best performance among the four proposed architectures.
In \texttt{dual--hum}, grammar scores were predicted from the final layer of the FFNN for grammatical features (see Figure~\ref{fig:model}), which might let the model learn better representations for grammatical features.

Since the \texttt{dual--hum} model performed the best, we conducted the subsequent experiments using the setting.

\subsection{Comparison of Grammatical Features}

\begin{table*}[t]
\centering
%\small
\tabcolsep 3pt
\begin{tabular}{lrrrrrrrrr}
\hline
 & \multicolumn{8}{l}{Prompt} & \\
\cline{2-10}
Features & 1 & 2 & 3 & 4 & 5 & 6 & 7 & 8 & avg. \\
\hline
baseline & .799\: & .662\: & .662\: & .804\: & .801\: & .809\: & .821\: & .726\: & .760\: \\
multi--ffnn1 & .803\: & .680\: & .659\: & .797\: & .802\: & .806\: & .827\: & .723\: & .762\: \\
multi--ffnn3 & .812\: & .671\: & .684\: & .812\: & .805\: & .812\: & .831\: & .748\: & .772\: \\
\hline
type256 & .818\: & .673\: & .687\: & .819\: & .807\: & .813\: & .833\: & .750\: & .775\: \\
\rowcolor{cyan!7}
\cellcolor[rgb]{1, 1, 1} & (\textbf{+ \!.019}) & (+ \!.011) & (\textbf{+ \!.025}) & (+ \!.015) & (+ \!.006) & (+ \!.004) & (\textbf{+ \!.012}) & (\textbf{+ \!.024}) & (+ \!.015) \\
err54   & .815\: & .672\: & .689\: & .813\: & .805\: & .812\: & .832\: & .756\: & .774\: \\
\rowcolor{cyan!7}
\cellcolor[rgb]{1, 1, 1} & (+ \!.016) & (+ \!.010) & (\textbf{+ \!.027}) & (+ \!.009) & (+ \!.004) & (+ \!.003) & (+ \!.011) & (\textbf{+ \!.030}) & (+ \!.014) \\
\hline
type256 \!+ \!err54 & .821\: & .673\: & .689\: & .815\: & .810\: & .805\: & .834\: & .752\: & .775\: \\
\rowcolor{cyan!7}
\cellcolor[rgb]{1, 1, 1} & (\textbf{+ \!.022}) & (+ \!.011) & (\textbf{+ \!.027}) & (+ \!.011) & (+ \!.009) & (- \!.004) & (\textbf{+ \!.013}) & (\textbf{+ \!.026}) & (+ \!.015) \\
\hline
multiply\_b & .811\: & .680\: & .701\: & .818\: & .813\: & .821\: & .829\: & .759\: & \textbf{.779}\: \\
\rowcolor{cyan!7}
\cellcolor[rgb]{1, 1, 1} & (+ \!.012) & (+ \!.018) & (\textbf{+ \!.039}) & (+ \!.014) & (+ \!.012) & (+ \!.012) & (+ \!.008) & (\textbf{+ \!.033}) & (+ \!.019) \\
prob & .820\: & .661\: & .682\: & .813\: & .807\: & .808\: & .834\: & .752\: & .772\: \\
\rowcolor{cyan!7}
\cellcolor[rgb]{1, 1, 1} & (\textbf{+ \!.021}) & (- \!.001) & (\textbf{+ \!.020}) & (+ \!.009) & (+ \!.006) & (- \!.001) & (\textbf{+ \!.013}) & (\textbf{+ \!.026}) & (+ \!.012) \\
multiply\_prob & .826\: & .662\: & .678\: & .815\: & .813\: & .809\: & .827\: & .746\: & .772\: \\
\rowcolor{cyan!7}
\cellcolor[rgb]{1, 1, 1} & (\textbf{+ \!.027}) & ($\pm$ \!0) & (+ \!.016) & (+ \!.011) & (+ \!.012) & ($\pm$ \!0) & (+ \!.006) & (\textbf{+ \!.020}) & (+ \!.012) \\
add\_prob & .812\: & .674\: & .682\: & .806\: & .799\: & .812\: & .827\: & .757\: & .771\: \\
\rowcolor{cyan!7}
\cellcolor[rgb]{1, 1, 1} & (+ \!.013) & (+ \!.012) & (\textbf{+ \!.020}) & (+ \!.002) & (- \!.002) & (+ \!.003) & (+ \!.006) & (\textbf{+ \!.031}) & (+ \!.011) \\
\hline
\hline
\citet{yang-etal-2020-enhancing} & .817\: & \textbf{.719}\: & .698\: & .845\: & \textbf{.841}\: & \textbf{.847}\: & .839\: & .744\: & \textbf{.794}\: \\
\rowcolor{cyan!7}
\cellcolor[rgb]{1, 1, 1} & (+ \!.017) & (\textbf{+ \!.040}) & (+ \!.019) & (\textbf{+ \!.023}) & (\textbf{+ \!.038}) & (\textbf{+ \!.050}) & (+ \!.004) & (+ \!.019) & (\textbf{+ \!.026}) \\
\citet{cao-2020-domain}          & .824\: & .699\: & \textbf{.726}\: & \textbf{.859}\: & .822\: & .828\: & \textbf{.840}\: & .726\: & .791\: \\
\rowcolor{cyan!7}
\cellcolor[rgb]{1, 1, 1} & (- \!.002) & (+ \!.001) & (+ \!.017) & (\textbf{+ \!.037}) & (- \!.002) & (- \!.001) & (+ \!.011) & (- \!.017) & (+ \!.006) \\
\citet{wang-etal-2022-use}       & \textbf{.834}\: & .716\: & .714\: & .812\: & .813\: & .836\: & .839\: & \textbf{.766}\: & .791\: \\
\hline
\end{tabular}
\caption{Comparison among grammatical features (dual--hum, QWK test). The numbers in parentheses indicate the improvement from the baseline. The numbers in parentheses for \citet{yang-etal-2020-enhancing} and \citet{cao-2020-domain} are the improvement from their baseline, which is equivalent to ours (\texttt{RegressionOnly} and \texttt{BERT \!(individual)}, respectively; n/a for \citet{wang-etal-2022-use}).}
\label{tab:dual-results}
\end{table*}

Using the \texttt{dual--hum} setting, we compared the effectiveness of different grammatical features.
Table~\ref{tab:dual-results} shows the QWK results on the test sets when we trained models using different grammatical features.\footnote{The QWK results for the auxiliary task are shown in Appendix~\ref{sec:appendix-aux-task}.}

\paragraph{PFs and NFs}
In the previous section, we showed that positive linguistic features (PFs; \texttt{type256}) improved the AES performance.
From the Table~\ref{tab:dual-results}, we can see that negative linguistic features (NFs; \texttt{err54}) also improved the model performance (see avg.).
Even on a per-prompt basis, the QWK scores were higher for all prompts than those in the baseline.

Combining the PFs and the NFs (\texttt{type256 \!\!+ \!\!err54}) also resulted in an improvement in AES performance.
However, the average QWK score (.775) was almost same as that for \texttt{type256} and \texttt{err54}, and no synergistic effect was observed by using both the PFs and the NFs.
We just concatenated the vectors of the two features before feeding the features to the FFNN for grammatical features, and there might be more effective ways to combine them.

\paragraph{PFs weighted using IRT parameters}
We further explored the effectiveness of PFs by weighting them using IRT parameters (see Section~\ref{sssec:weighted-pf}).
When we considered the difficulties of individual grammatical items (\texttt{multiply\_b}), the QWK score became the highest among all settings.
On the other hand, modifying \texttt{type256} with the probability that a writer with a certain grammar ability uses the grammatical item correctly did not help to improve AES performance.
Although the QWK scores got higher than that for the baseline, they were lower than that for \texttt{type256}.
The results suggest that it is more important to capture what items the writer actually used in the essay than what items the writer seemed able to use.

\paragraph{Effect of Grammatical Features}
To verify that the score improvement came from the addition of grammatical features rather than multi-task learning, we trained models with the auxiliary task but without using grammatical features.
The number of hidden layers in the FFNN on the top was set to 1 (\texttt{multi--ffnn1}; same as the baseline) and 3 (\texttt{multi--ffnn3}; the best parameter for \texttt{multi--hum}; see Section~\ref{ssec:hyperparameters-for-models}).
The QWK scores for \texttt{multi--ffnn1} and \texttt{multi--ffnn3} were higher than that of the baseline, but lower than those of the models with grammatical features (Table~\ref{tab:dual-results}).
The results show that both multi-task learning and grammatical features contributed to improve the model performance.
In addition, the significant improvement on \texttt{multi--ffnn3} suggests that adding layers on the top of BERT would be effective in multi-task learning. 

\paragraph{Scoring examples}
We show some examples from the fold~2 of Prompt~1 (Table~\ref{tab:scoring-example}).
The true scores of the four examples are 10, and are written in roughly the same number of words.

In ID~1382, a relatively wide variety of grammatical items were used (10.18 items per 100 words, while the average for essays with true score of 10 included in the fold~2 test set was 9.86).
The model trained using \texttt{type256} captured the characteristic and predicted the correct score.

On the other hand, for ID~377 and 104, the model trained using \texttt{type256} assigned lower scores than the baseline because of the limited variety of grammatical items in the essays.
Note that the prediction improved in ID~377, while it got worse in ID~104.

In ID~1097, our model did not perform well.
Although a relatively wide variety of grammatical items were used, the predicted score was lower than that of the baseline.\footnote{See Appendix~\ref{sec:appendix-confusion-matrix} for the confusion matrix on all the data points.}

\begin{table}[t]
\centering
\small
\tabcolsep 3pt
\begin{tabular}{rrrrrr}
\hline
 & & \multicolumn{2}{l}{Grammatical items} & \multicolumn{2}{l}{Predicted score} \\
\cline{3-6}
Essay ID & \# words & \# type & per 100 & baseline & type256 \\
\hline
1382 & 442 & 45 & 10.18 & 9 & 10 \\
377 & 480 & 47 & 9.79 & 12 & 11 \\
104 & 405 & 38 & 9.38 & 9 & 8 \\
1097 & 421 & 42 & 9.98 & 9 & 8 \\
\hline
\end{tabular}
\caption{Scoring examples. The true scores of the four examples are 10. Per 100 represents the number of different grammar items used per 100 words.}
\label{tab:scoring-example}
\end{table}

\paragraph{Comparison with existing models}
The QWK scores for the state-of-the-art AES models are also shown in Table~\ref{tab:dual-results}.
The average QWK score of our models (the highest at .779) was not as high as those of the existing models.
In some prompts, there seemed to be differences in baseline QWK scores between the previous studies and ours, and we made comparisons based on the improvement from each baseline\footnote{We re-implemented $\text{R}^2$ BERT \citep{yang-etal-2020-enhancing}, but our re-implementation of the model did not achieve as good scores as those reported in their paper. Furthermore, we trained models using grammatical features with the loss combination proposed by them (\emph{i.e.}, regression and ranking loss), which resulted in lower QWK scores than our baseline.}.

In Prompt 1, 3, 7, and 8, our proposed models showed a greater improvement in the QWK scores compared to \citet{yang-etal-2020-enhancing} and \citet{cao-2020-domain}.
In these four prompts, the scores themselves of our models were also competitive with those of the existing models.
\citet{cao-2020-domain} achieved the state-of-the-art results in Prompt~3, 4, and 7, but the improvements from their baselines were relatively small in the other prompts.

However, our proposed methods were less effective for Prompt 2, 4, 5, and 6, which resulted in lower average QWK scores than the existing models.
To identify when the proposed methods were effective, we examined the characteristics of the essays, such as the type of essays, the average number of words in essays, the correlation coefficient between holistic scores and the grammar ability parameter $\theta$ and between human-annotated grammar scores and $\theta$, and the variance of $\theta$, but none of them could provide a satisfactory explanation.
We need further investigation and it might help to improve the performance on the prompts where our methods were less effective.

\subsection{Using the IRT Ability Parameter for the Auxiliary Task}

\begin{table}[t]
\centering
\small
\tabcolsep 2pt
\begin{tabular}{lccccccccc}
\hline
 & \multicolumn{8}{l}{Prompt} & \\
\cline{2-10}
Model & 1 & 2 & 3 & 4 & 5 & 6 & 7 & 8 & avg. \\
\hline
multi-irt & .819 & .669 & .697 & .811 & .813 & .821 & .839 & .757 & .778 \\
dual-irt  & .805 & .678 & .686 & .807 & .808 & .816 & .831 & .742 & .772 \\
\hline
\end{tabular}
\caption{QWK results of the models using the IRT ability parameter for the auxiliary task (QWK test)}
\label{tab:irt-multi-dual}
\end{table}

In Section~\ref{ssec:compare-model}, we demonstrated that \texttt{dual--hum} model achieved the best performance among the four proposed architectures. 
However, the architecture requires grammar scores annotated by human raters.
Therefore we employed grammar abilities estimated using IRT, which requires no human-annotated labels, as the teacher signals.

Table~\ref{tab:irt-multi-dual} shows that \texttt{multi--irt} and \texttt{dual--irt} models achieved comparable performance to the models that used human-annotated score.
In general, analytical scoring is more time-consuming than holistic scoring, and grammar scores, which are one of the analytical scores, are not always available in a dataset.
A method that improves AES performance without the additional human-annotated labels has practical value.
Another advantage of using IRT for our AES models is that we can provide the characteristics of grammatical items (\emph{i.e.}, discrimination and difficulty) as well as essay scores.

\section{Conclusions}

This study examined the effectiveness of using grammatical features in AES models.
Specifically, we fed two kinds of features: (1)~grammatical items that writers used correctly in essays (PFs), and (2)~the number of grammatical errors (NFs).
We showed that both PFs and NFs improved the model performance, but combining them did not result in further improvement.
The experimental results suggest that multi-task learning would be effective to take advantage of the information that the grammatical features have.
One of the future directions could be exploring effective ways to combine PFs and NFs to improve the model performance since the way in this study was a simple concatenation of the two vectors (\emph{e.g.}, to learn representations for PFs and NFs in different networks and combine them).
Another direction would be to examine the effectiveness of adding our grammatical features in AES using a large language model.
It potentially improves the scoring performance in zero- and/or few-shot settings \cite{mizumoto-2023}.
Furthermore, in order to have more interpretable models, it would be beneficial to analyze how much individual grammatical features contribute to model's score prediction.
The insights delivered by interpretable models can help practitioners in education.

We also weighted PFs in several ways using IRT parameters and found that considering the difficulties of grammatical items would improve the model performance.
In addition, we used the ability parameter $\theta$ as teacher signals for the auxiliary task in multi-task learning.
Although no human-annotated labels are required to estimate the IRT parameters, the model trained with the ability parameter achieved comparable performance to the model trained with grammar scores annotated by human raters.
In this study, IRT parameters were estimated based on grammatical items that writers used in their essays.
In the future, we will apply IRT to both PFs and NFs to model writers' grammar abilities.

\section{Limitations}

Our proposed methods showed significant advantage on some essay prompts in the ASAP dataset, while they were less effective on the other prompts.
Further investigation is necessary to clarify what kind of essays our proposed methods would be effective to.
An analysis of the effectiveness of grammatical features on different prompts will also provide additional insights into the variation of model behavior across different prompts.

There are also some limitations related to the extraction of grammatical features.
First, the toolkit provided by the CEFR-J Grammar Profile extracts grammatical items based on sophisticated regular expression patterns, which was written by a linguist.
It would be quite challenging to prepare a similar toolkit in other languages.
\citet{banno-matassoni-2022-cross} let a model predict the frequencies of grammatical errors from essay representations, which can be applicable to PFs, but the approach requires human-annotated labels to train a model.
Another approach is to extract grammatical features based on cross-linguistically consistent annotations such as Universal Dependencies.
It makes easier to use grammatical features in other languages, while it remains challenging to extract ones related to parts of speech and/or morphological features rather than dependencies (\emph{e.g.}, present perfect in English).

Second, there could be errors in the extraction using regular expressions and the same is true for grammatical error correction.
Experiments using grammatical features annotated by humans would help reveal the influence of errors in feature extraction.

Third, our method requires explicitly extracting grammatical features at test time as well as at training time.
An alternative would be to develop a multi-task learning framework where a model is trained to reconstruct grammatical features at training time and then run the trained model on unparsed test data (\emph{e.g.}, \citealp{andersenbenefits-2021}).

\section*{Acknowledgments}
A part of this work was supported by JSPS KAKENHI Grant Numbers JP21H05054 and by JST, the establishment of university fellowships towards the creation of science technology innovation, Grant Number JPMJFS2137.

%%%%%%%%%%%%%%%%%%%%%%%%%%

\bibliography{anthology,custom}
% \bibliography{custom}

%\clearpage
\appendix

\begin{table*}[t!]
\centering
%\small
\tabcolsep 2pt
\begin{tabular}{lccccccccc}
\hline
 & \multicolumn{8}{l}{Prompt} & \\
\cline{2-10}
Model & 1 & 2 & 3 & 4 & 5 & 6 & 7 & 8 & avg. \\
\hline
type256 & 0.032 & -0.007 & 0.050 & -0.007 & 0.001 & 0.000 & 0.000 & 0.079 & 0.016 \\
err54 & -0.002 & 0.017 & -0.003 & 0.014 & -0.007 & 0.003 & -0.012 & 0.045 & 0.008 \\
type256+err54 & 0.148 & 0.003 & 0.085 & 0.001 & 0.000 & 0.001 & -0.002 & 0.110 & 0.039 \\
\hline
multiply\_b & 0.015 & -0.003 & -0.002 & -0.023 & 0.000 & 0.000 & 0.012 & -0.003 & -0.001 \\
prob & 0.052 & -0.025 & 0.028 & 0.000 & 0.000 & 0.000 & 0.000 & 0.046 & 0.008 \\
multiply\_prob & 0.097 & 0.000 & 0.003 & 0.000 & 0.000 & 0.000 & 0.007 & 0.059 & 0.018 \\
add\_prob & 0.053 & -0.023 & 0.050 & -0.002 & 0.004 & 0.000 & 0.039 & 0.004 & 0.011 \\
\hline
\end{tabular}
\caption{QWK results for the auxiliary task on the test set (models shown in Table~\ref{tab:dual-results})}
\label{tab:qwk-aux}
\end{table*}

\section{QWK Results for the Auxiliary Task}
\label{sec:appendix-aux-task}

Table~\ref{tab:qwk-aux} shows the QWK score for the auxiliary task (\emph{i.e.}, predicting grammar score).
The QWK scores were generally low, and some of them were negative.
We observed that the models output scores close to the mode value in each prompt.
One of the possible reasons is the relatively low weight for loss function for the auxiliary task (\emph{i.e.}, 0.2).
However, when we assigned a higher weight for the auxiliary task (\emph{i.e.}, 0.4), the model prediction for the main task got worse.
Further consideration is necessary for predicting multiple essay traits simultaneously (\emph{e.g.}, \citealp{Ridley_He_Dai_Huang_Chen_2021, shibata-uto-2022-analytic}).

\section{Detailed Results of Model Predictions}
\label{sec:appendix-confusion-matrix}

Detailed scoring performance of the model trained using \texttt{type256} is shown in Figure~\ref{fig:confusion-matrix}.
The values in the confusion matrices are the sum of all experiments (\emph{i.e.}, 5-fold cross validation and three experiments with different seed values).

\setcounter{figure}{1}
\begin{figure*}[ht]
\includegraphics[width=0.5\linewidth]{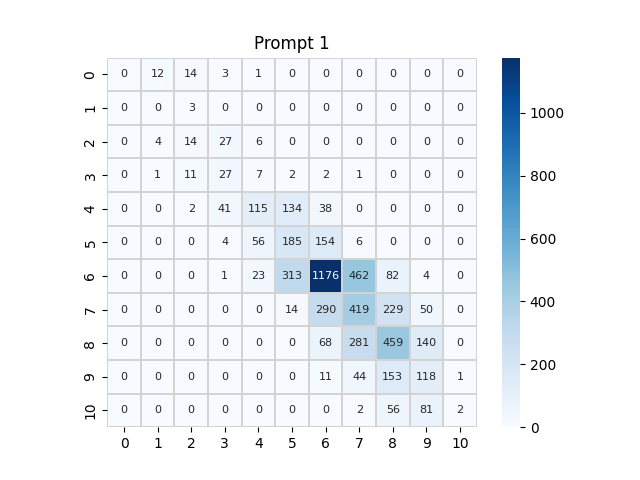} \hfill
\includegraphics[width=0.5\linewidth]{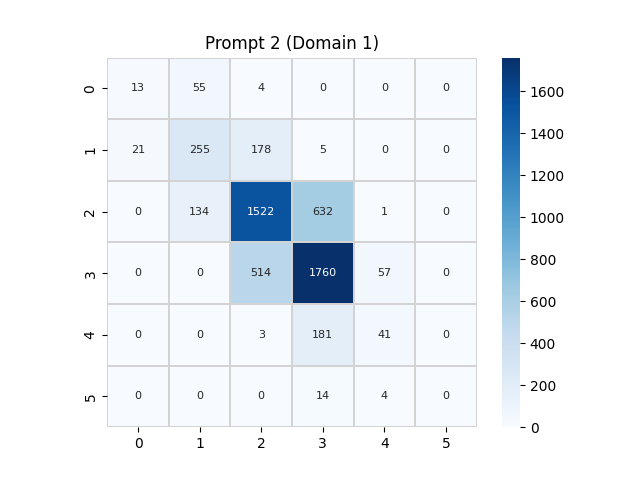} \hfill
\includegraphics[width=0.5\linewidth]{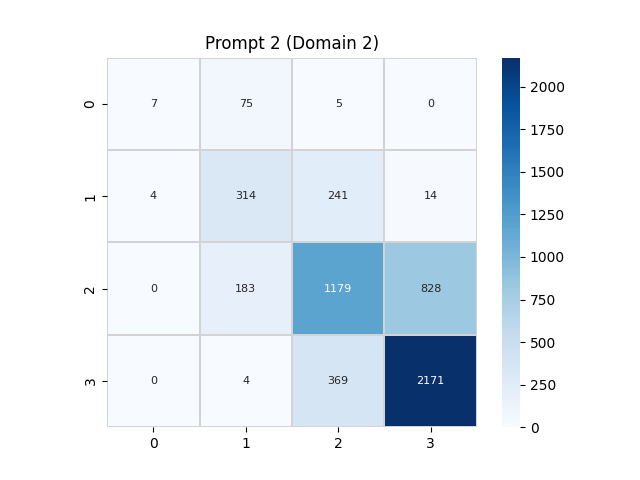} \hfill
\includegraphics[width=0.5\linewidth]{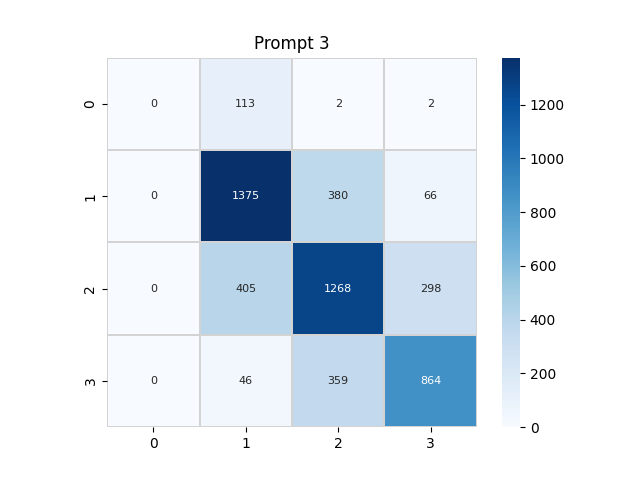} \hfill
\includegraphics[width=0.5\linewidth]{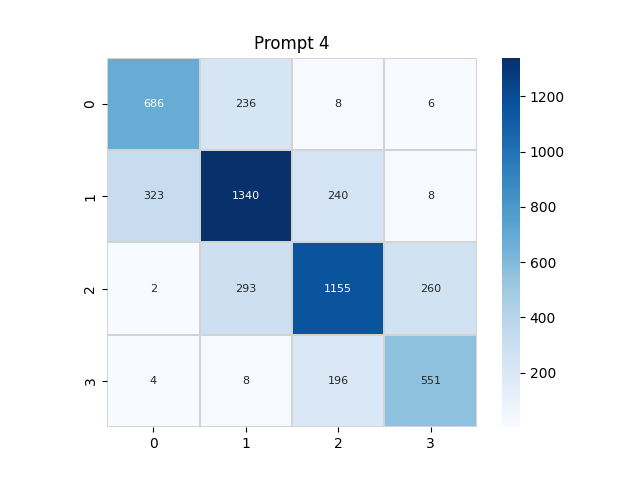} \hfill
\includegraphics[width=0.5\linewidth]{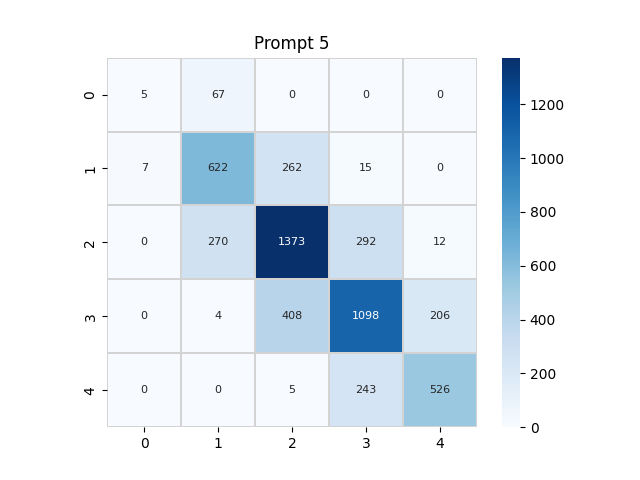} \hfill
\caption {Scoring performance of the model trained using \texttt{type256}}
\label{fig:confusion-matrix}
\end{figure*}

\begin{figure*}[ht]
\includegraphics[width=0.5\linewidth]{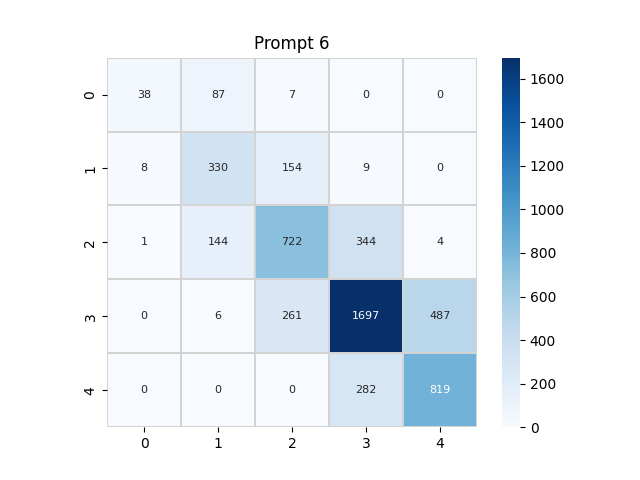} \hfill
\includegraphics[width=0.5\linewidth]{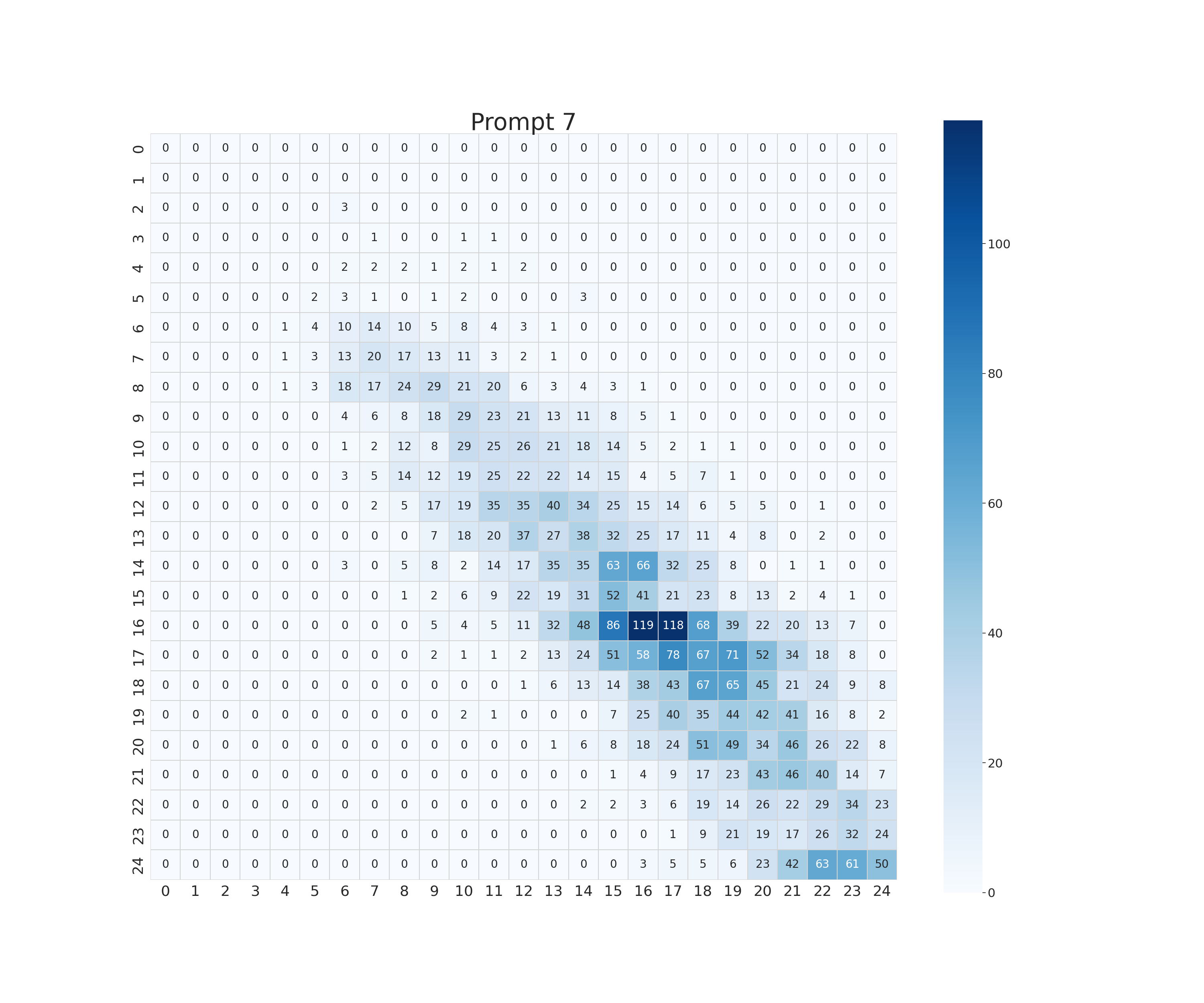} \hfill
%\caption {Scoring performance of the model trained using \texttt{type256} (\emph{cont.})}
\end{figure*}

\setcounter{figure}{1}
\begin{figure*}[t]
\hspace{-1.0cm}
\includegraphics[width=18.0cm]{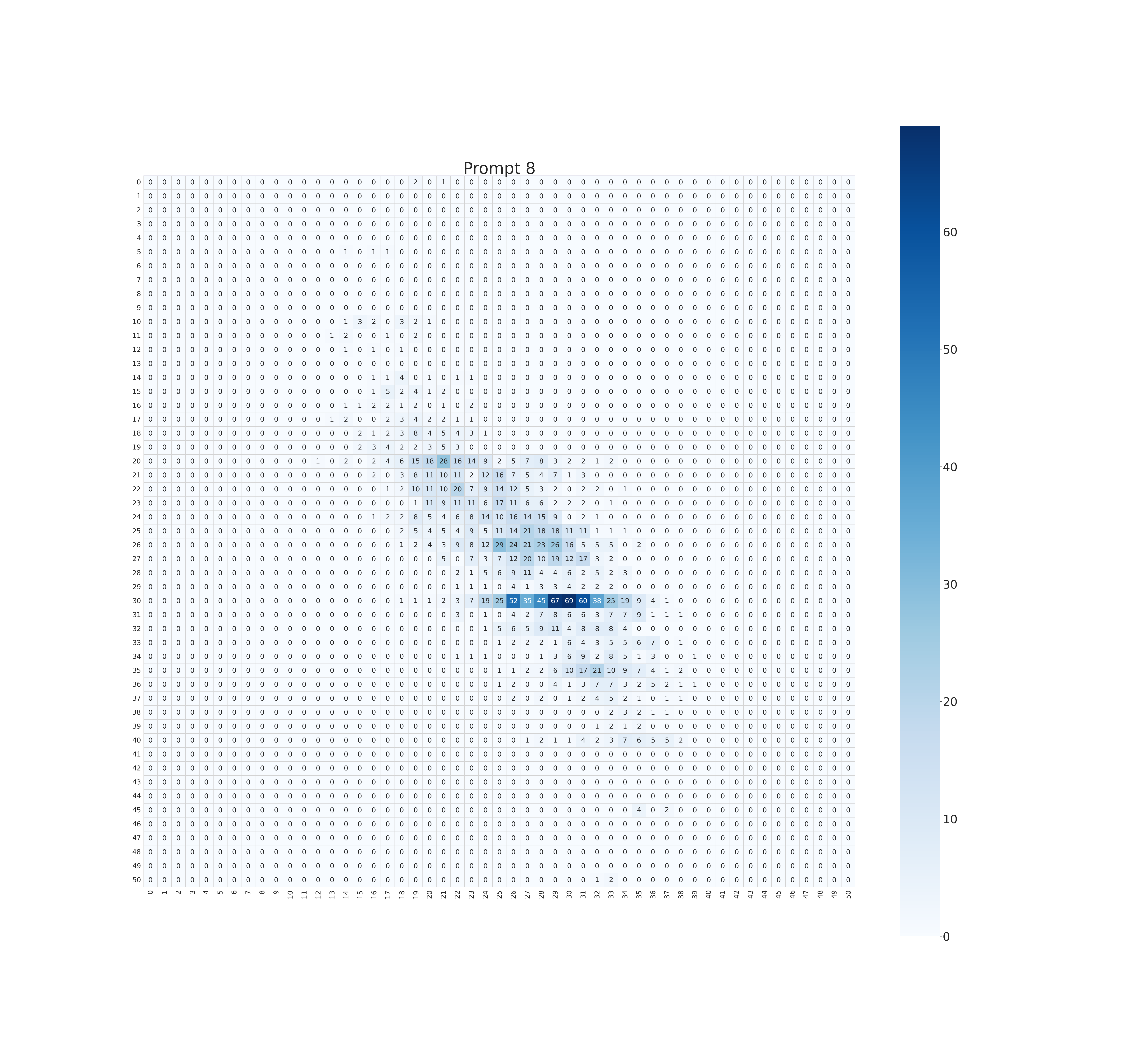}
\caption {Scoring performance of the model trained using \texttt{type256} (\emph{cont.})}
\end{figure*}

\end{document}